\documentclass[journal,10pt]{IEEEtran}
\usepackage{mathrsfs}
\usepackage{cite}
\usepackage{epsfig}
\usepackage{epsf}
\usepackage{graphics}
\usepackage{subfig}
\usepackage{algorithm}
\usepackage{algorithmicx}
\usepackage{mathtools}
\usepackage{algpseudocode}
\usepackage{graphicx}

\usepackage{color}
\usepackage{url}
\usepackage{hyperref}
\usepackage{flushend}
\usepackage{epstopdf}
\usepackage{cite}
\usepackage{graphicx}
\usepackage{textcomp}
\usepackage{bm}
\usepackage{xcolor}
\usepackage{tabularx}
\def\L{\bm{l}}

\def\PTM{\text{MTEC}}

\def\W{\mathit{W}}
\def\V{\bm{V}}
\def\Q{\bm{Q}}
\def\K{\bm{K}}
\def\Z{\bm{Z}}

\def\E{\bm{E}}

\def\R{\mathbb{R}}

\def\d{d}

\def\dh{d_h}
\def\l{l}
\def\h{h}

\def\L{L}
\def\BibTeX{{\rm B\kern-.05em{\sc i\kern-.025em b}\kern-.08em
    T\kern-.1667em\lower.7ex\hbox{E}\kern-.125emX}}
\usepackage{amsmath,graphicx,amssymb,amsfonts}
\usepackage{algorithm}
\usepackage{algpseudocode}
\def\BibTeX{{\rm B\kern-.05em{\sc i\kern-.025em b}\kern-.08em
    T\kern-.1667em\lower.7ex\hbox{E}\kern-.125emX}}
\definecolor{brown}{rgb}{0.59, 0.29, 0.0}
\begin{document}
\title{Multi-Content Time-Series Popularity Prediction with Multiple-Model Transformers in MEC Networks}

\author{Zohreh HajiAkhondi-Meybodi,~\IEEEmembership{Student Member,~IEEE}, Arash Mohammadi,~\IEEEmembership{Senior Member,~IEEE}, Ming Hou,~\IEEEmembership{Senior Member,~IEEE}, Elahe Rahimian,~\IEEEmembership{Student Member,~IEEE}, Shahin Heidarian,~\IEEEmembership{Student Member,~IEEE}, 
Jamshid Abouei,~\IEEEmembership{Senior Member,~IEEE}, and~Konstantinos~N.~Plataniotis,~\IEEEmembership{Fellow,~IEEE}% <-this % stops a space
\thanks{Z. HajiAkhondi-Meybodi is with Electrical and Computer Engineering (ECE), Concordia University, Montreal, Canada. (E-mail: z\_hajiak@encs.concordia.ca). A. Mohammadi (corresponding author) is with Concordia Institute of Information Systems Engineering (CIISE), Concordia University, Montreal, Canada. (P: +1 (514) 848-2424 ext. 2712 F: +1 (514) 848-3171, E-mail: arash.mohammadi@concordia.ca).  Ming Hou is with Defence Research and Development Canada (DRDC), Ottawa, Toronto, ON,
M2K 3C9, Canada. (E-mail: ming.hou@drdc-rddc.gc.ca). E. Rahimian is with Concordia Institute of Information Systems Engineering (CIISE), Concordia University, Montreal, Canada. (E-mail: e\_ahimia@encs.concordia.ca). S. Heidarian is with Electrical and Computer Engineering (ECE), Concordia University, Montreal, Canada. (E-mail: shahin.heidarian@concordia.ca). J. Abouei was with the Department of Electrical and Computer Engineering, University of Toronto, Toronto, Canada. He is now with the Department of Electrical Engineering, Yazd University, Yazd 89195-741, Iran (E-mail: abouei@yazd.ac.ir). K.~N.~Plataniotis is with Electrical and Computer Engineering (ECE), University of Toronto, Toronto, Canada. (E-mail: kostas@ece.utoronto.ca).}

\thanks{This Project was partially supported by Department of National Defence's Innovation for Defence Excellence \& Security (IDEaS)
program, Canada.}}

\markboth{Internet of Things Journal,~Vol.~XX, No.~X, XX~2022}%
{Hajiakhondi \MakeLowercase{\textit{et al.}}:Multi-Content Time-Series Popularity Prediction with Multiple-Model Transformers in MEC Networks}
\maketitle
\begin{abstract}
Coded/uncoded content placement in Mobile Edge Caching (MEC) has evolved as an efficient solution to meet the significant growth of global mobile data traffic by boosting the content diversity in the storage of caching nodes. To meet the dynamic nature of the historical request pattern of multimedia contents, the main focus of recent researches has been shifted to develop data-driven and real-time caching schemes. In this regard and with the assumption that users' preferences remain unchanged over a short horizon, the Top-$K$ popular contents are identified as the output of the learning model. Most existing data-driven popularity prediction models, however, are not suitable for the coded/uncoded content placement frameworks. On the one hand, in coded/uncoded content placement, in addition to classifying contents into two groups, i.e., popular and non-popular, the probability of content request is required to identify which content should be stored partially/completely, where this information is not provided by existing data-driven popularity prediction models. On the other hand, the assumption that users' preferences remain unchanged  over a short horizon only works for content with a smooth request pattern. To tackle these challenges, we develop a Multiple-model (hybrid) Transformer-based Edge Caching ($\PTM$) framework with higher generalization ability, suitable for various types of content with different time-varying behavior, that can be adapted with coded/uncoded content placement frameworks. Simulation results corroborate the effectiveness of the proposed $\PTM$ caching framework in comparison to its counterparts in terms of the cache-hit ratio, classification accuracy, and the transferred byte volume.
\end{abstract}

\IEEEpeerreviewmaketitle

%OOOOOOOOOOOOOOOOOOOOOOOOOOOOOOOOOOOOOOOOOOOOOOOOOOOOOOOO
\section{Introduction} \label{Sec:1}
%OOOOOOOOOOOOOOOOOOOOOOOOOOOOOOOOOOOOOOOOOOOOOOOOOOOOOOOO
Mobile Edge Caching (MEC)~\cite{Liu2021, Abbas2018, Khan2020} has emerged as a promising solution for potential deployment in the Sixth Generation ($6$G) of communication networks to meet the phenomenal growth of global mobile data traffic. The main idea behind MEC networks is to provide low-latency communication for Internet of Things (IoT) devices by bringing multimedia content closer to users~\cite{Hajiakhondi2019, Hajiakhondi2020}. In this context, if content requested by an IoT device can be found in the storage of one of the nearby caching nodes, low-latency communication will be established and cache-hit occurs, otherwise, the IoT device will experience high latency. Due to the limited storage of caching nodes, one of the efficient approaches to improve the cache-hit ratio is to increase the content diversity. This can be fulfilled by implementing a coded/uncoded content placement in an integrated MEC network~\cite{Chen2017_2, Lin2020_1}. Recent advancements in heterogeneous cluster-centric cellular networks~\cite{Lin2019} have drawn focused research attention given provided considerable improvements in content diversity, which is due to the integration of the coded/uncoded content placement and Coordinated Multi-Point (CoMP) technology. Besides, integrating Unmanned Aerial Vehicles (UAVs) as flying caching nodes into the cluster-centric MEC networks~\cite{Hajiakhondi2021} can further improve the network's Quality of Service (QoS) due to the high-quality Line of Sight (LoS) links, high mobility of UAVs, and their wide transmission range.

Despite all the research works conducted to develop an efficient coded/uncoded content placement strategy, the dynamic and time-varying topology of MEC networks presents new challenges when it comes to the design of an optimal real-time caching scheme~\cite{Dai2022}. On the one hand, integrated MEC networks are unpredictable in real-world scenarios due to the high mobility of both mobile users and edge caching devices~\cite{Zhang2022}. Additionally, mobile users' preferences are time-varying, depending on different variables such as content popularity, geographical region, and the users' contextual information~\cite{Hajiakhondi2021_ICC}. On the other hand, the local storage capacity of cache-enabled edge devices is limited, therefore, it is essential to dynamically observe the users' request pattern to regularly update the storage of edge devices with the most upcoming popular contents. To accommodate these critical aspects of MEC networks, it is of significant practical importance to augment MEC networks with a data-driven popularity prediction model to continuously monitor and analyse time-varying request patterns of multimedia contents. The paper aims to further advance this emerging field.

%OOOOOOOOOOOOOOOOOOOOOOOOOOOOOOOOOOOOOOOOOOOOOOOOOOOOOOOOO
\noindent
\textbf{Literature Review:}
%OOOOOOOOOOOOOOOOOOOOOOOOOOOOOOOOOOOOOOOOOOOOOOOOOOOOOOOOO
Recently, several promising approaches have been developed to predict the popularity of multimedia content, including but not limited to $(i)$ Statistical models~\cite{Marlin2011, Odic2013}, such as  collaborative filtering, item-to-item correlation systems, and content-based filtering; $(ii)$ Machine Learning (ML)-based architectures~\cite{Abidi2020}, such as Generalized Linear Model (GLM)~\cite{Ng2019}, Decision Tree (DT)~\cite{Kabra2011}, and Random Forest (RF)~\cite{Mendez2008}, and; $(iii)$ Deep Neural Networks (DNN)~\cite{Doan2018, Ale2019, Fan2021, Zhang2019, Rathore2019, Lin2020, Zhong2020, Wu2019, Wang2019}, such as Convolutional Neural Network (CNN)~\cite{Tsai2018}, and Long Short Term Memory (LSTM)~\cite{Zhang2019, Mou2019}. Despite all the benefits that come from existing statistical and ML-based frameworks, they suffer from sparsity and cold-start problems, which arise when sufficient information is not provided about a new mobile user/multimedia content. Moreover, having a well-trained ML architecture relies on an efficient feature engineering model to extract several contextual information associated with both mobile users and multimedia content. While ML-based caching schemes suffer from poor scalability across different scenarios, DNN-based  models can capture the user's interests from raw historical request patterns without the need for feature extraction and pre-processing. Consequently, the main focus of recent research has been shifted to DNN-based frameworks to monitor/predict content popularity using the historical request pattern of contents.

With the focus on DNN models, Yu~\textit{et al.}~\cite{Yu2021} introduced an auto-encoder architecture to learn the latent representation of historical request patterns of content to predict the users' preferences in the upcoming time. Although auto-encoder is an unsupervised (self-supervised) learning model and there is no need for ground truth labels of content popularity, it suffers from several drawbacks. On the one hand, auto-encoder needs a large number of training data, resulting in a high training overhead, i.e., excessive training complexity. On the other hand, auto-encoder attempts to capture as much information as possible rather than capturing the relevant information. Therefore, in a scenario where the relevant information is just a small part of the data, the model fails to train well. Ndikumana~\textit{et al.}~\cite{Ndikumana2021} used CNN architecture to capture the contextual information of users, such as age, emotion, and gender to calculate the probability of requesting a content using Multi-Layer Perceptron (MLP). To have an efficient DNN-based caching scheme, however, both temporal and spatial correlations of content should be captured from time-variant and sequential request patterns of multimedia content. More precisely, spatial correlation represents different users' preferences, relying on regional information, content popularity, and users' contextual information. The temporal correlation is related to the time-variant behavior of the request pattern. While CNN-based caching schemes can capture local spatial correlations, they fail to properly capture temporal features of historical request patterns. Furthermore, such models require a data pre-processing stage to provide some additional information to be used as the input features and improve the cache performance.

To address the above-mentioned challenges associated with CNN-based caching schemes, Recurrent Neural Networks (RNNs), such as LSTM~\cite{Zhang2019, Mou2019}, were introduced to capture the temporal dependencies of sequential request patterns. While LSTM models have advantages for processing time-series data, their computation complexity, unsuitability for capturing long-term dependencies, and difficulties for parallel computations make them problematic. To tackle the aforementioned issues, Transformer architectures~\cite{Vaswani2017} have been developed. Similar to RNN models, Transformers are designed to process sequential data, while there is no need to analyze the sequential data in the same order, resulting in higher parallelization, and reduced training complexity. In~\cite{Hajiakhondi2021_ICC}, we introduced a Vision Transformer (ViT)-based Edge (TEDGE) caching framework for an uncoded content placement, which to the best of our knowledge, was being studied for the first time to predict the Top-$K$ popular content in MEC networks. Our prior work~\cite{Hajiakhondi2021_ICC} and other existing data-driven popularity prediction models are, however, unsuitable for coded/uncoded content placement frameworks. The reason is that the existing classification models classify content into two groups, i.e., popular and non-popular, in which popular contents are stored completely. In a coded/uncoded content placement, however, multimedia content should be stored completely/partially based on its popularity, i.e., the request probability. In addition, the assumption that users' preferences remain unchanged over a short time horizon, only works for contents with smooth request patterns. The paper addresses these gaps as outlined below.

%OOOOOOOOOOOOOOOOOOOOOOOOOOOOOOOOOOOOOOOOOOOOOOOOOOOOOOOOO
\vspace{.025in}
\noindent
\textbf{Contributions:}  To tackle the aforementioned challenges, in this paper, we develop the Multiple-model (hybrid) Transformer-based Edge Caching ($\PTM$) framework as a multi-content and time-series popularity prediction model. The $\PTM$ framework captures both temporal and spatial correlation of multiple contents via multi-channel Transformer architectures, where the sequential request patterns of each content are given to a channel of the Transformer model. Our first objective for the development of the $\PTM$ framework is to introduce a data-driven popularity prediction model with higher generalizability compared to existing  works that predict the Top-$K$ popular content relying on the historical request pattern. The second objective is to adapt the data-driven prediction model within the coded/uncoded content placement approaches.  To achieve these objectives, the proposed $\PTM$ framework is built upon the Transformer architecture, consisting of two parallel paths, which takes the historical request pattern of multiple contents as its input:
\begin{itemize}
\item The first path of the $\PTM$ framework is a Transformer network, responsible for identifying the Top-$K$ popular contents in the upcoming time, using the historical request pattern of contents. This part of the architecture is efficient for contents that their request patterns are smoothly changed over a short horizon.
\item The second path of the $\PTM$ framework is introduced to boost the generalizability of the learning model. In this path, for applicability to various types of content with different time-varying behavior, we relax the assumption of unchanged request patterns of content. Moreover, within the context of coded/uncoded content placement, we take one step forward and relax the assumption that the probability of content requests is known a-priori (i.e., Zipf distribution)~\cite{Hajiakhondi2022}. In this regard, the second path is compromised of two stages, where the first stage is a Transformer network responsible for predicting the probability of requesting multiple contents in the upcoming time. The estimated request probability as the output of the first stage will be used for the coded/uncoded content placement to identify which content should be stored partially/completely in the storage of caching nodes. Next, it will be concatenated with the historical request pattern (input of the first stage), provided as the input to the second stage, which analyzes the popularity of all contents simultaneously.
\item The final output, which is the combination of the two parallel paths, is the Top-$K$ popular contents in the upcoming time, which is applicable to various types of content with different time-varying behavior. The effectiveness of the proposed $\PTM$ framework is evaluated through comprehensive studies on the real-trace multimedia request pattern, in terms of the classification accuracy, cache-hit ratio, and the transferred byte volume. Simulation results corroborate the effectiveness of the proposed $\PTM$ framework in comparison to its counterparts over all the aforementioned aspects.
\end{itemize}

The remainder of the paper is organized as follows: In Section~\ref{Sec:2}, the system model is described and the main assumptions required for the implementation of the proposed framework are introduced. Section~\ref{Sec:3} presents the proposed MTEC scheme. Simulation results are presented in Section~\ref{Sec:4}. Finally, Section~\ref{Sec:5} concludes the paper.

%%%%%%%%%%%%%%%%%%%%%%%%%%%%%%%%%%%%%%%%%%
\begin{figure} [t!]
\centering \includegraphics [scale = 0.38] {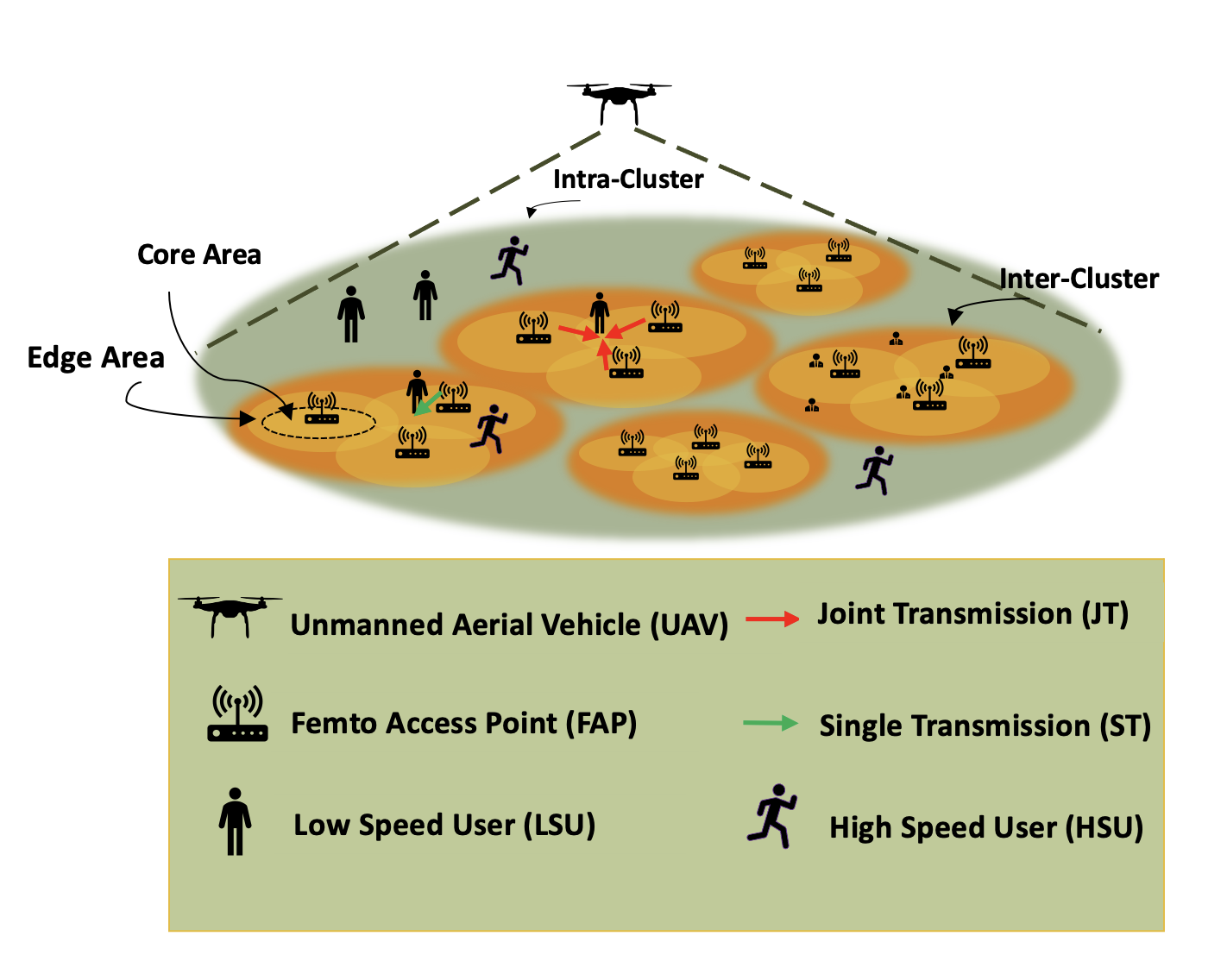}
\vspace{-.1in}
\caption{\footnotesize A typical structure of a cluster-centric and UAV-aided cellular networks.} \label{sys}
\vspace{-.1in}
\end{figure}
%%%%%%%%%%%%%%%%%%%%%%%%%%%%%%%%%%%%%%%%%%

%OOOOOOOOOOOOOOOOOOOOOOOOOOOOOOOOOOOOOOOOOOOOOOOOOOOOOOOO
\section{System Model and Problem Description} \label{Sec:2}
%OOOOOOOOOOOOOOOOOOOOOOOOOOOOOOOOOOOOOOOOOOOOOOOOOOOOOOOO
We consider a cluster-centric UAV-aided cellular network in both indoor and outdoor residential areas. As shown in Fig.~\ref{sys}, there are $N_f$ number of FAPs, denoted by $f_i$, for ($1 \leq i \leq N_f$), distributed based on the Poisson Point Processes (PPPs) in the environment~\cite{Chen2017_2}, $N_u$ number of UAVs, denoted by $u_{k}$, for ($1 \leq k \leq N_u$), and the main server. Followed by Gaussian mixture distribution, there exist $N_g$ number of ground users, denoted by $GU_j$, for ($1 \leq j \leq N_g$), moving through the network with the velocity of $\upsilon_{j}(t)$. To avoid frequent handover in terrestrial infrastructure and UAV's signal attenuation in indoor areas, indoor requests are managed by FAPs, while outdoor users' requests are handled through FAPs/UAVs, depending on their movement speed (i.e., Low-Speed Users (LSU) are served through FAPs, otherwise, they are managed by UAVs)~\cite{Hajiakhondi2022}. Due to the high mobility of users, the location of UAVs is determined using the $K$-means clustering algorithm~\cite{Hajiakhondi2021}. Each UAV covers one intra-cluster and hovers at its location while delivering a request~\cite{Fadlullah2020}. Besides, in contrast to conventional femtocaching schemes, where each FAP acts as a single unity, we assume that the storage of $N_b < N_f$ number of nearby FAPs belonging to an inter-cluster is known as a component. 

%OOOOOOOOOOOOOOOOOOOOOOOOOOOOOOOOOOOOOOOOOOOOOOOOOOOOOOOO
%\subsection{Coded/Uncoded Content Placement}\label{SubSec:CP}
%OOOOOOOOOOOOOOOOOOOOOOOOOOOOOOOOOOOOOOOOOOOOOOOOOOOOOOOO
There is a content library $\mathcal{C} = \lbrace c_1, \ldots, c_{N_c} \rbrace$, where  $N_c=|\mathcal{C}|$ is the cardinality of contents in the network, and each content  $c_l$ is segmented into $N_s$ encoded parts, denoted by $c_{ls}$, for ($ 1\leq s \leq N_s$). While the storage capacity of FAPs and UAVs, denoted by $C_f$ and  $C_u$, respectively, is limited, the main server has access to the whole library of multimedia content. Multimedia contents are conventionally classified using the Zipf distribution~\cite{Chen2017_2} into three groups, i.e., popular, mediocre, and non-popular. After sorting content based on their popularity in the descending order followed by the Zipf distribution, it is assumed that $\alpha$ percent of FAPs' cache capacity is assigned to complete popular content, i.e., $1 \leq l \leq \lfloor \alpha C_f \rfloor$, while the remaining part is allocated to  different parts of the mediocre content $c_l$, where $\lfloor \alpha C_f \rfloor +1 \leq l \leq  N_s ( C_f - \lfloor \alpha C_f \rfloor)$. 

To deliver multimedia content from inter-clusters to users, two transmission schemes, i.e., Single Transmission (ST) and Joint Transmission (JT), are utilized based on the link quality of the user and the popularity of the requested content~\cite{Hajiakhondi2022}. More precisely, if the average Signal-to-Interference-plus-Noise Ratio (SINR) of user  $GU_j$ and FAP $f_i$, denoted by $\overline{\mathcal{S}}_{i,j}(t)$, is higher than a pre-defined threshold $\mathcal{S}_{th}$, then user $GU_j$ is known as the cell-core of FAP $f_i$.  In this case and regardless of the popularity of the content, this request will be managed by FAP $f_i$, i.e., the ST scheme. On the other hand, if the requested content is popular and the user $GU_j$ is marked as the cell-edge of FAP $f_i$, this request will be jointly served by all FAPs belonging to the corresponding inter-cluster, i.e., the JT scheme. This completes the description of the system model, which is used after predicting the content popularity by the proposed $\PTM$ multiple-model (hybrid) Transformer architecture. Next, we introduce our proposed MTEC framework.

%OOOOOOOOOOOOOOOOOOOOOOOOOOOOOOOOOOOOOOOOOOOOOOOOOOOOOOOO
\section{Multiple-Model Transformer-based Edge Caching ($\PTM$) Framework} \label{Sec:3}
%OOOOOOOOOOOOOOOOOOOOOOOOOOOOOOOOOOOOOOOOOOOOOOOOOOOOOOOO
The main goal of the proposed $\PTM$ architecture is to predict the Top-$K$ popular content using the historical request pattern of the underlying contents. In this context, we use MovieLens Dataset~\cite{Harper2015}, where leaving a comment after watching a movie is considered a request~\cite{Zhang2019, Dernbach2016, Li2016}. In this section, we first describe the dataset pre-processing phase, and then present different blocks of the proposed $\PTM$ architecture.
%%%%%%%%%%%%%%%%%%%%%%%%%%%%%%%%%%%%%%%%
\begin{figure*}[t!]
\centering
\includegraphics[scale=0.32]{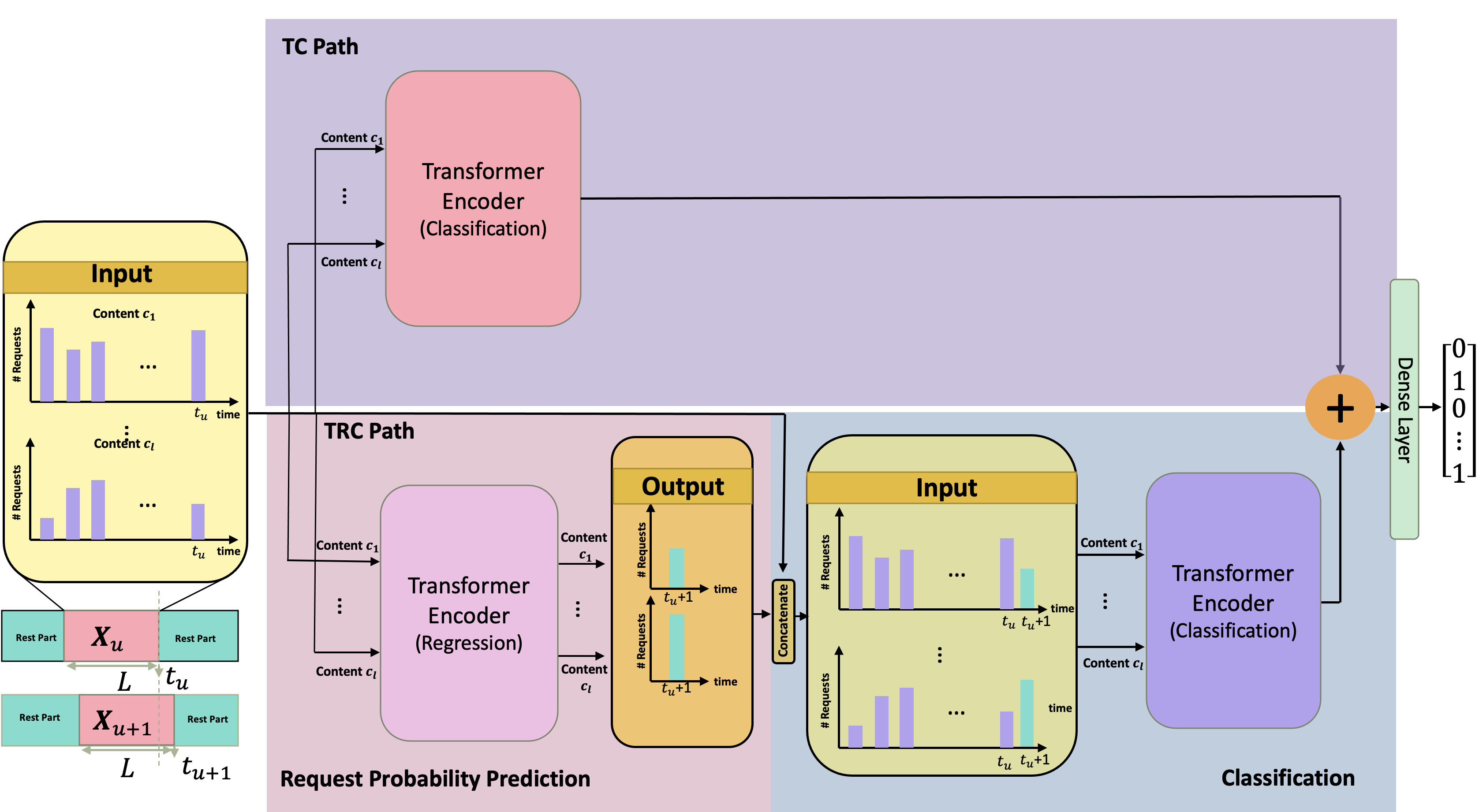}
\caption{Block diagram of the proposed $\PTM$ architecture.}\label{BlockDiagram}
\end{figure*}
%%%%%%%%%%%%%%%%%%%%%%%%%%%%%%%%%%%%%%%%

%OOOOOOOOOOOOOOOOOOOOOOOOOOOOOOOOOOOOOOOOOOOOOOOOOOOOOOOO
\subsection{Dataset Pre-processing} \label{pre}
%OOOOOOOOOOOOOOOOOOOOOOOOOOOOOOOOOOOOOOOOOOOOOOOOOOOOOOOO
MovieLens Dataset includes users' contextual information, such as gender, age, and occupation together with their geographical information, i.e., ZIP code. To determine users' location during their requests, we convert ZIP codes to longitude and latitude coordinates~\cite{Zhang2019}. Given the caching nodes' location and their transmission range, therefore, a set of caching nodes in the vicinity of each user will be determined. To adopt MovieLens Dataset to our popularity prediction model, we perform the following four steps:

\noindent
\textit{\textbf{Step 1 - Request Matrix Formation}}: Since the proposed $\PTM$ multiple-model Transformer architecture is a time-series forecasting model, we first sort the requests of content $c_l$, for ($1 \leq l \leq N_c$), in the ascending order of time. With the assumption that there are $T$ timestamps (i.e., seconds) and $N_c$ number of contents, $\textbf{R} \in \R^{N_c \times T}$ represents an indicator request matrix for each caching node as follows
\begin{eqnarray}
\textbf{R} =
\begin{bmatrix}
1 &  \ldots  & 0 \\
0 &  \ldots  & 1 \\
\vdots & \ddots & \vdots \\
1 & \ldots & 0
\end{bmatrix}_{N_c \times T}
\end{eqnarray}
where $r_{l,t}=1$ means that content $c_l$ is requested at time $t$; otherwise, $r_{l,t}=0$.

\noindent
\textit{\textbf{Step 2 - Time Windowing}}:  Considering the fact that content placement will be performed during the off-peak time (updating time $t_u$)~\cite{Vallero2020}, it is assumed that time $T$ is discretized into $N_{w}$ number of time intervals with a length of $\mathcal{W}$, where $\mathcal{W}$ represents the time duration between two off-peak times. Consequently, we form a window-based request matrix, denoted by $\textbf{R}^{(\mathcal{W})} \in \R^{ N_c  \times N_{w}}$, where $r^{(w)}_{l,t_u}= \sum \limits_{t=(t_u-1)\mathcal{W}+1}^{t_u\mathcal{W}} r_{l,t}$ is the cumulative requests of content $c_l$ during two consecutive updating times $t_u-1$ and $t_u$.

\noindent
\textit{\textbf{Step 3 - Data Segmentation}}: To predict the Top-$K$ popular content at updating time $t_u$, the input data should be a historical request pattern of content during the past updating times with the length of $L$. Given $\textbf{R}^{(\mathcal{W})}$ from the previous step, the window-based request matrix $\textbf{R}^{(\mathcal{W})}$ is segmented via an overlapping sliding window of length $L$ to provide input samples $\textbf{X}_u \in \mathbb{R}^{N_c \times M}$, where $M = \dfrac{N_{w}}{L}$ is the total number of segments (input samples). We, therefore, have  $\mathcal{D}=\{(\textbf{X}_u, \textbf{y}_u)\}_{u=1}^{M}$, with $\textbf{y}_u \in  \mathbb{R}^{N_c \times 1}$ as the output (label) of the learning model, where $\sum \limits_{l=1}^{N_c} y_{u,l} = K$, and $y_{u,l}=1$ means that content $c_l$ would be popular at $t_{u+1}$, otherwise, it would be zero.

%%%%%%%%%%%%%%%%%%%%%%%%%%%%%%%%%%%%%%%%
\setlength{\textfloatsep}{0pt}
\begin{figure*}[h]
\centering
\includegraphics[scale=0.3]{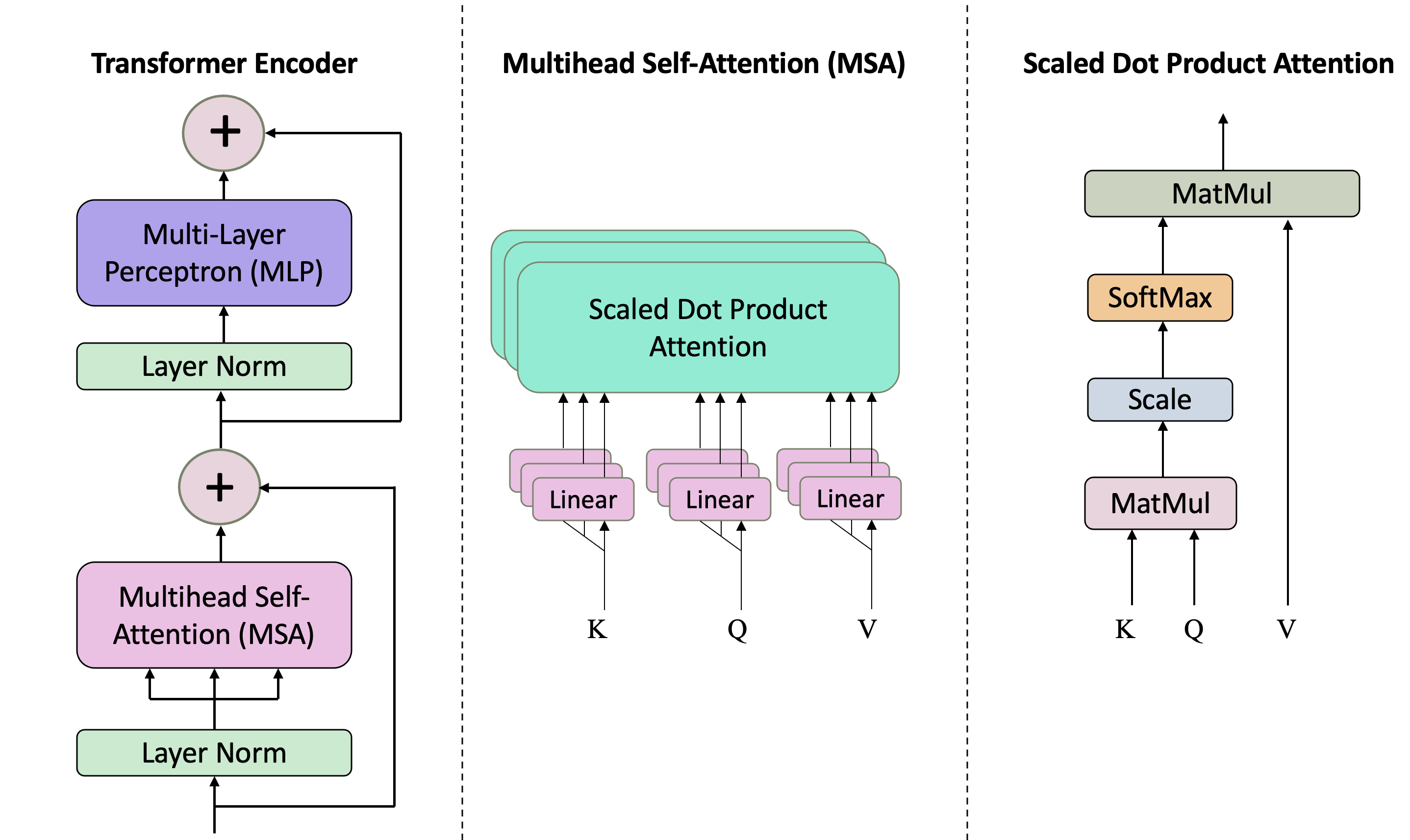}
\caption{Left: Architecture of the Transformer encoder, Middle: Multihead Self-Attention (MSA); Right: Scaled dot product attention.}\label{transformer}
\end{figure*}
%%%%%%%%%%%%%%%%%%%%%%%%%%%%%%%%%%%%%%%%
\noindent
\textit{\textbf{Step 4 - Data Labeling}}: Relying on the historical request pattern of content at updating time $t_u$, denoted by $\textbf{X}_u$, our goal is to predict the  Top-$K$ popular content, which is a multi-class classification problem. To identify (label) content as popular/non-popular ones, we have the following two criteria:
\begin{itemize}
\item[$i)$] \textit{Probability of Requesting a Specific Content}: Probability of requesting $c_l$, for ($1 \leq l \leq N_c$), at updating time $t_u$, denoted by $p_l^{(t_u)}$, is given by
\begin{equation}
p_l^{(t_u)}=\dfrac{r^{(w)}_{l,t_u}}{\sum \limits_{l=1}^{N_c}r^{(w)}_{l,t_u}},
\end{equation}
where $r^{(w)}_{l,t_u}$ represents the number of requests of content $c_l$ in time window with length of $L$.
\item[$ii)$] \textit{Skewness of the Request Pattern}, which is a commonly used metric in time-series forecasting models to accelerate the popularity prediction of the first appearance content~\cite{Joseph2001}. Term $\zeta_l$ represents the skewness of content $c_l$, where negative skew means that the number of requests of content $c_l$ increases over time. Therefore,  the Top-$K$ content $c_l$, for ($1 \leq l \leq K$), with negative skew and the highest probability, will be labeled with $y_{u,l}=1$ and identified as the Top-$K$ popular content.
\end{itemize}
%

%AC: This section is short in length, shallow, but given that it is the main section of the paper, it should be much more detailed. One suggestion is to bring the related material from Section II and discuss them here as part of discussing your mode. For example, start with the good introduction you have here, then move on to dtaset discussion, transformer model should be discussed in Section II.A; discussion on content placement should come to the subsection on Request Probability Prediction Block. and then at the end bring the loss function information.

%OOOOOOOOOOOOOOOOOOOOOOOOOOOOOOOOOOOOOOOOOOOOOOOOOOOOOOOO
\subsection{MTEC Architecture} \label{MTEC}
%OOOOOOOOOOOOOOOOOOOOOOOOOOOOOOOOOOOOOOOOOOOOOOOOOOOOOOOO
In this subsection, we present the constituent components of the proposed $\PTM$ framework, where the main architecture is developed based on the Transformers (see Fig.~\ref{BlockDiagram}). There are following drawbacks to the existing research works that motivate us to develop the $\PTM$ framework:

%In our previous work~\cite{Hajiakhondi2021_ICC}, we introduced the TEDGE caching scheme, as a multi-label classification model, developed based on a single ViT architecture. 
%
\begin{itemize}
\item[$i)$] With the assumption that users’ preferences remain unchanged  over a short horizon, existing works~\cite{Hajiakhondi2021_ICC, Ndikumana2021, Tsai2018} predicted the Top-$K$ popular content in the updating time $t_{u+1}$, where the input of the learning model was the request pattern of all contents in a time window with the length of $L$, ending at the updating time of $t_u$. This assumption works for content whose request pattern smoothly changes over time. It is, therefore, essential to develop a popularity prediction model with higher generalization ability, suitable for various types of content with different time-varying behavior.
\item[$ii)$] In a coded/uncoded content placement, the request probability of content in an upcoming time is also required to classify the Top-$K$ popular content into two groups, i.e., popular and mediocre, while existing classification frameworks~\cite{Hajiakhondi2021_ICC, Mou2019, Yu2021, Ndikumana2021} classified content as popular/non-popular.
\end{itemize}
To tackle the aforementioned challenges, we propose the $\PTM$ framework, consisting of two parallel paths, where the first path is a Transformer-based Classification (TC) model, and the second path includes two series of Transformer-based blocks, named Transformer-based Regression and Classification (TRC) network. These are followed by a fully connected layer, as a fusion center combining the output of the two parallel paths to estimate the Top-$K$ popular content. It should be noted that although the output of these two paths is similar in nature, simulation results illustrate that considering such an architecture improves the popularity prediction. Next, we introduce each of these blocks.

%=========================================================
\subsubsection{\textbf{TC Path}}
%=========================================================
This path is a multi-label classification model based on the Transformer model, where the input is the historical request pattern of multiple contents at time $t_u$ and the output is the Top-$K$ popular content at time $t_{u+1}$. The Transformer is a type of Machine Learning (ML) model suitable for learning from sequential and time series data. Generally speaking, Transformers outperform the LSTM models because: $(i)$ Although LSTM models are capable of learning long-term dependencies, they suffer from short-term memory over long sequences. Transformers, however, capture the connection/dependency between sequential components that are far from one another, resulting in higher accuracy; $(ii)$ Due to the Multi-head Self-Attention (MSA) mechanism, Transformers can process data in parallel, reducing the training time; and, $(iii)$ The attention mechanism of the Transformer eliminates the need to analyze data in the same order. Consequently, positional embedding is used to preserve the position information of an entity in sequential data.

The multi-content training sample $\textbf{X}_u \in \mathbb{R}^{N_c \times M}$ includes $M$ feature vectors $\textbf{x}_u \in \mathbb{R}^{N_c \times 1}$. Using the min-max normalization method, the feature vectors are first normalized and then linearly projected
into a $d$-dimensional vector space, where $d$ represents the model dimension. The input of the learning model $\textbf{v}_u \in \mathbb{R}^{d}$ is given by
\begin{eqnarray}
\textbf{v}_u = \textbf{W}_p \textbf{x}_u + \textbf{b}_p,
\end{eqnarray}
where $\textbf{W}_p \in \mathbb{R}^{d \times N_c}$ and $\textbf{b}_p \in \mathbb{R}^{d}$ are learnable parameters. Followed by Reference~\cite{Zerveas2021} and to preserve the temporal correlations, we use a $1$D-convolutional layer with $d$ output channels to build vector $\textbf{v}_u$. The positional embedding $\E^{pos} \in \mathbb{R}^{M \times d}$ is then appended to the input vector $\textbf{V} \in \mathbb{R}^{M \times d} = [\textbf{v}_1, \ldots, \textbf{v}_{M}]$, to form the input of the transformer encoder $\Z_0 = \textbf{V} + \E^{pos}$~\cite{Vaswani2017}.

\vspace{.1in}
\noindent
\textit{\textbf{Transformer Encoder:}} As shown in Fig.~\ref{transformer}, there are $\L$ layers in transformer encoder, consisting of MSA and the MLP modules, as follows
\begin{eqnarray}
\Z^{'}_l &=& MSA(LayerNorm(\Z_{\l-1})) + \Z_{\l-1},\label{eq:MSA}\\
\Z_l &=& MLP(LayerNorm(\Z^{'}_{\l})) + \Z^{'}_{\l}, \label{eq:MLP}
\end{eqnarray}
where $\Z^{'}_l$ and $\Z_l$ represent the output of the MSA and MLP modules associated with layer $l$, for ($1 \leq l \leq L$), respectively. Note that, before applying MSA and MLP modules, we use a layer-normalization to address the degradation problem~\cite{layernorm}. Moreover, the Gaussian Error Linear Unit (GELU) is used as the activation function in the MLP module with two linear layers. This completes the description of the Transformer encoder developed for the design of the $\PTM$ architecture. Considering the fact that the MSA module is defined based on the Self-Attention (SA) mechanism, next, we present the description of SA and MSA modules.

\vspace{.1in}
\noindent
\textit{\textbf{1. Self-Attention (SA):}} The SA module~\cite{Vaswani2017} in the Transformer architecture is used to capture the dependency of different vectors in $\Z \in \R^{M \times d}$, where $\Z$ consists of $M$ vectors, each with size of $d$. In this regard, Query $\Q$, Key $\K$, and Value $\V$ matrices are calculated as follows
\begin{eqnarray}
[\Q, \K, \V] = \Z\W^{QKV}\label{eq.2},
\end{eqnarray}
where $\W^{QKV} \in \R^{\d\times 3\dh}$ is a trainable weight matrix and $\dh$ represents the dimension of $\Q$, $\K$, and $\V$ matrices. The output of the SA block  $SA(\Z) \in \R^{M \times \dh}$ is calculated as a weighted sum of the values $\V$, with the weights assigned to each value, determined by a compatibility function between the query and the relevant key, as follows
\begin{eqnarray}
SA(\Z) = \text{softmax} \left( \frac{\Q\K^T}{\sqrt{\dh}} \right)  \V\label{eq.3},
\end{eqnarray}
where the term $\dfrac{\Q\K^T}{\sqrt{\dh}}$ is the scaled dot-product of $\Q$ and $\K$ by $\sqrt{\dh}$, and $\text{softmax}$ is used to convert it to the probability values ranged between zero and one.

\vspace{.1in}
\noindent
\textit{\textbf{2. Multihead Self-Attention (MSA):}} The MSA module comprises of $\h$ heads with different trainable weight matrix $\{\W^{QKV}_i\}^{\h}_{i=1}$, performed $\h$ times in parallel. Given the outputs of $\h$ SA modules, the output of the MSA module is given by
\begin{eqnarray}
MSA(\Z) = [SA_1(\Z); SA_2(\Z); \dots; SA_h(\Z)]W^{MSA},
\end{eqnarray}
where $\W^{MSA} \in \R^{\h \dh \times \d}$ and $\dh$ is set to $\d / \h$. 

Although this block is capable of capturing the request pattern of content with predictable behavior, i.e., with smooth changes over time in uncoded content placement, it would not be effective for content with sudden changes in a coded/uncoded manner. For these reasons, the second path is required.

%%%%%%%%%%%%%%%%%%%%%%%%%%%%%%%%%%%%%%%%%%
\begin{figure} [th]
\centering \includegraphics [scale = 0.5] {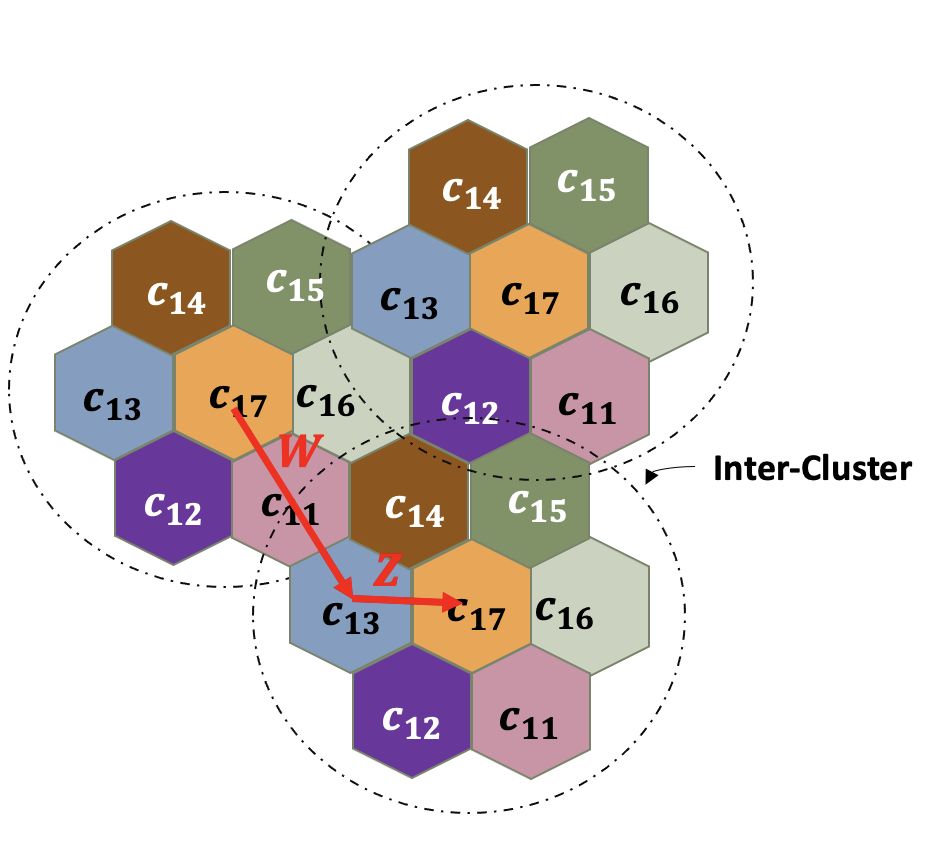}
%\vspace{-.1in}
\caption{\footnotesize Coded content placement scheme associated with mediocre content in single/multiple inter-clusters.} \label{content}
%\vspace{-.1in}
\end{figure}
%%%%%%%%%%%%%%%%%%%%%%%%%%%%%%%%%%%%%%%%%%

%=========================================================
\subsubsection{\textbf{TRC Path}}
The second path consists of the following two blocks:
\begin{itemize}
\item[$i)$] \textbf{Request Probability Prediction Block: } The first block of the second path is used to predict the request probability of content at time $t_{u+1}$ using the historical request pattern of content at time $t_u$ (input data is the same as the first path). The output of this block will be used to classify the Top-$K$ popular content (the final output of the $\PTM$ framework) as the popular/mediocre one in the coded/uncoded content placement. As discussed in Section~\ref{Sec:2}, the Top-$K$ popular content will be sorted in descending order, where $N_p=\lfloor \alpha C_f \rfloor $ and  $N_a= N_s ( C_f - \lfloor \alpha C_f \rfloor)$ are the cardinality of popular and mediocre content, respectively. Following our prior work~\cite{Hajiakhondi2022}, vector $\mathbf{z}=[z_1, \ldots, z_{N_a}]^T$ represents mediocre contents, where $N_a$ denotes the cardinality of mediocre content. To identify which segments of mediocre content $c_l$, denoted by $c_{ls}$ for $(1 \leq l \leq N_a)$ and $(1 \leq s \leq N_s)$, is cached in each FAP $f_i$ for $(1 \leq i \leq N_b)$ belonging to an inter-cluster, an indicator matrix  $\bm{Z}^{(f_i)}$ associated with FAP $f_i$ is formed, where the $l^{\text{th}}$ row of $\bm{Z}^{(f_i)}$, represented by $\bm{z}^{(f_i)}_{l}=[0,\ldots,0,  1]_{(1 \times N_s)}$ is corresponding to the segments of file $c_l$ cached in FAP $f_i$. Note that, $\sum \limits_{s=1}^{N_s} z^{(f_i)}_{ls} = 1$, where $z^{(f_i)}_{ls} = 1$, if the $s^{\text{th}}$ segment of file $c_l$ is cached in FAP $f_i$, otherwise, it would be zero. Therefore, the mediocre content of other FAPs $f_j$, for ($1 \leq j \leq N_b, j \neq i$), belonging to an inter-cluster, is given by
\begin{equation}\label{nEq:22}
\bm{z}^{(f_i)}_{l} {\bm{z}_{l}^{(f_j)}}^{T}=0,~~~~i=1,\ldots,N_b, j=1,\ldots,N_b, i\neq j.
\end{equation}
Following the above discussion, different segments of mediocre content will be stored in an inter-cluster. Next, the same content as FAP $f_i$ in one inter-cluster, is allocated to FAP $f_k$ in the nearby inter-cluster, where $k$ is given by
\begin{equation}\label{nEq:23}
\bm{Z}^{(f_{k})}= \bm{Z}^{(f_i)} ~~~ \text{if}~~~ k=w^2+wz+z^2,
\end{equation}
where $w$ is the number of FAPs required to move from FAP $f_i$ in any direction, after which $z$ number of FAPs should be moved by turning $60$ degrees counterclockwise to reach FAP $f_k$~\cite{Hajiakhondi2022} (see Fig.~\ref{content}). Please refer to our previous  work~\cite{Hajiakhondi2022} for a detailed description of the coded/uncoded content placement. Finally, the estimated request probability is appended to the original input samples to generate the input of the next block, which is used for the classification, i.e., identifying the Top-$K$ popular content.

\item[$ii)$] \textbf{Classification Block: } In comparison to the CT block in the first path, the input of this block is both the historical request pattern of content at time $t_u$ and the estimated one at $t_{u+1}$, resulting in higher classification accuracy for such content with sharp changes in their request pattern. Then, the output features of both paths are added, which are used as the input of the fusion layer (dense layer). The output of the dense layer is a vector $\textbf{y}_u \in \R^{N_c}$, with $K$ ones, where $1^{'}$s indicates the Top-$K$ popular content (i.e., popular/mediocre one) and  $0^{'}$s are non-popular content. Finally, the estimated Top-$K$ popular contents are categorized into popular/mediocre one according to the output of the request probability prediction block.
\end{itemize}

Finally, we use Mean Squared Error (MSE) for the request probability prediction block, and binary cross-entropy as the loss function for the CT path, the classification block in the second path, and the fusion path. The overall loss function $\mathcal{L}$ of the proposed MTEC is given by
\begin{equation}\label{loss}
\mathcal{L} = w_1 \mathcal{L}_{RPP} + w_2 \mathcal{L}_{C_{I}} + w_3 \mathcal{L}_{C_{II}} + w_4 \mathcal{L}_{F},
\end{equation}
where $\mathcal{L}_{RPP}$, $\mathcal{L}_{C_{I}}$, $\mathcal{L}_{C_{II}}$, and $\mathcal{L}_{F}$ represent the loss function associated with the request probability prediction block, CT block, the classification in the second path, and the fusion path, respectively, and $w_i$, $i=\{1,\ldots,4\}$ is the loss weight of each block.

%OOOOOOOOOOOOOOOOOOOOOOOOOOOOOOOOOOOOOOOOOOOOOOOOOOOOOOOO
\section{Simulation Results} \label{Sec:4}
%OOOOOOOOOOOOOOOOOOOOOOOOOOOOOOOOOOOOOOOOOOOOOOOOOOOOOOOO
To evaluate the performance of the proposed $\PTM$ framework, we consider a cluster-centric UAV-aided cellular network with $138,493$ GUs and $27,278$ number of distinct contents. Following the common assumption~\cite{Hajiakhondi2019}, we consider the scenario that the storage capacity of caching nodes is $10\%$ of the total content, where the size of all multimedia contents is the same.
%AC: Can you add a reference to then state that following the common assumption [?], we consider the scenario that the storage ...
Considering the GUs' location, determined based on their ZIP code~\cite{Ndikumana2021}, it is assumed that there are $21$, and $3$ terrestrial, and areal caching nodes, respectively, where each inter-cluster consists of $N_s=7$ number of FAPs. Note that the classification accuracy is averaged over all caching nodes.  %The general simulation parameters of the cluster-centric UAV-aided cellular network are followed by our previous work~\cite{Hajiakhondi2022}.
%AC: Possibly add a table with the parameters information to make this paper self-contained OK
To determine the best architecture of the multiple-model Transformer-based architecture, we first evaluate different versions of the proposed $\PTM$ popularity prediction framework through trial and error. Moreover, the performance of a single classification model, i.e., Path 1 or Path 2, is evaluated, when they are trained independently. In all the experiments, the Adam optimizer is employed with learning rate of $0.0001$ and weight decay of $0.00001$. The activation function of the MLP layers in all Transformer models is ReLU, while it is sigmoid as the output layer.  
In classification blocks, the multi-content time-series request pattern data is converted to a sequential set of images, using the Gramian Angular Field (GAF) technique~\cite{Hong2020} to preserve the temporal correlations of the input data. 
%Finally, we use Mean Squared Error (MSE) for the Request Probability Prediction block, and binary cross-entropy as the loss function for the Classification I and II, and the fusion path. The overall loss function $\mathcal{L}$ of the proposed MTEC is given by
%
%\begin{equation}\label{loss}
%\mathcal{L} = w_1 \mathcal{L}_{RPP} + w_2 \mathcal{L}_{C_{I}} + w_3 \mathcal{L}_{C_{II}} + w_4 \mathcal{L}_{F},
%\end{equation}
%
%where $\mathcal{L}_{RPP}$, $\mathcal{L}_{C_{I}}$, $\mathcal{L}_{C_{II}}$, and $\mathcal{L}_{F}$ represent the loss function associated with the Request Probability Prediction block, Classification I, Classification II, and Fusion path, respectively, and $w_i$, $i=\{1,\ldots,4\}$ is the loss weight of each block.
%AC: The above discussion on loss function can possibly be  moved to the model section.

%%%%%%%%%%%%%%%%%%%%%%%%%%%%%%%%%%%%%%%%
\begin{table*}[t]
\centering
\renewcommand\arraystretch{2}
\caption{\small Variants of the $\PTM$ popularity prediction framework.}
\label{table1}
{\begin{tabular}{   c | c c c  c  c  c | c }
\hline
\hline
\textbf{Model ID}
& \textbf{Layers (L)}
& \textbf{Model dimension ($\d$)}
& \textbf{MLP layers}
& \textbf{MLP size}
& \textbf{Number of Heads}
& \textbf{Params}
& \textbf{Accuracy}
\\
\hline
\textbf{1}
& 1
& 32
& 1
& 256
& 8
& 444,073
&  84.32 $\%$
\\
\textbf{2}
& 1
& 32
& 1
& 256
& 16
& 652,457
& 88.15 $\%$
\\
\textbf{3}
& 1
& 64
& 1
& 256
& 8
& 775,049
& 88.23 $\%$
\\
 \textbf{4}
& 2
& 64
& 1
& 256
& 8
& 1,438,556
& 92.78 $\%$
\\
\textbf{5}
& 1
& 64
& 1
& 512
& 8
& 885,384
& 90.01 $\%$
\\
\textbf{6}
& 2
& 64
& 2
& 256
& 16
& 2,882,140
& 94.13 $\%$
\\
\hline
\end{tabular}}
\end{table*}
%%%%%%%%%%%%%%%%%%%%%%%%%%%%%%%%%%%%%%%%

%%%%%%%%%%%%%%%%%%%%%%%%%%%%%%%%%%%%%%%%
\begin{table}[t]
\centering
\renewcommand\arraystretch{2}
\caption{\small The accuracy of the Top-$K$ popular content using different window size (9, 49, and 99 days) for different variants of the proposed MTEC framework.}
\label{table2}
{\begin{tabular}{   c | c  c  c  }
\hline
\textbf{Model ID}  &  & \textbf{Accuracy} & \\
&\textbf{9 Days}
& \textbf{49 Days}
& \textbf{99 Days}
\\
\hline
\textbf{1}
&  81.09 $\%$
&  84.32 $\%$
&  85.14 $\%$
\\
\textbf{2}
& 86.09 $\%$
& 88.15 $\%$
& 88.86 $\%$
\\
\textbf{3}
& 85.94 $\%$
& 88.23 $\%$
& 89.21 $\%$
\\
 \textbf{4}
& 89.45 $\%$
& 92.78 $\%$
& 93.03 $\%$
\\
\textbf{5}
& 87.54 $\%$
& 90.01  $\%$
& 90.98 $\%$
\\
\textbf{6}
& 92.87 $\%$
& 94.13  $\%$
& 94.54 $\%$
\\
\hline
\end{tabular}}
\end{table}
%%%%%%%%%%%%%%%%%%%%%%%%%%%%%%%%%%%%%%%%

%%%%%%%%%%%%%%%%%%%%%%%%%%%%%%%%%%%%%%%%
\begin{table}[t]
\centering
\renewcommand\arraystretch{2}
\caption{\small The accuracy of the Top-$K$ popular content using different loss weights.}
\label{table3}
{\begin{tabular}{   c | c  c  c  c | c}
\hline
\textbf{Model ID} 
&\textbf{$w_1$}
& \textbf{$w_2$}
& \textbf{$w_3$}
& \textbf{$w_4$}
& \textbf{Accuracy}
\\
\hline
     \textbf{L1}
& 0.2
& 0.4
& 0.1
& 0.3
& 94.13 $\%$
\\
    \textbf{L2}
& 0.3
& 0.2
& 0.1
& 0.4
& 93.08 $\%$
\\
\textbf{L3}
& 0.0
& 0.0
& 0.0
& 1.0
& 90.54 $\%$
\\
\hline
\end{tabular}}
\end{table}
%%%%%%%%%%%%%%%%%%%%%%%%%%%%%%%%%%%%%%%%

%%%%%%%%%%%%%%%%%%%%%%%%%%%%%%%%%%%%%%%%
\begin{table*}[t]
\centering
\renewcommand\arraystretch{2}
\caption{\small Comparison between our methodology ($\PTM$ framework) and other versions of the Transformer-based architectures.}
\label{table4}
{\begin{tabular}{   c | c c c  c  c  c | c }
\hline
\hline
\textbf{Model Name}
& \textbf{Layers (L)}
& \textbf{Model dimension ($\d$)}
& \textbf{MLP layers}
& \textbf{MLP size}
& \textbf{Number of Heads}
& \textbf{Parameters}
& \textbf{Accuracy}
\\
\hline
     \textbf{TC}
& 2
& 64
& 2
& 256
& 16
& 946,580
& 92.48 $\%$
\\
\textbf{RCT}
& 2
& 64
& 2
& 256
& 16
& 1,909,860
& 92.54 $\%$
\\
\textbf{ViT}
& 2
& 64
& 2
& 256
& 16
& 3,798,320
& 92.68 $\%$
\\
 \textbf{$\PTM$}
& 2
& 64
& 2
& 256
& 16
& 2,882,140
& 94.13 $\%$
\\
\textbf{$\PTM$-AT}
& 2
& 64
& 2
& 256
& 16
& 2,882,396
& 93.28 $\%$
\\
\hline
\end{tabular}}
\end{table*}
%%%%%%%%%%%%%%%%%%%%%%%%%%%%%%%%%%%%%%%%

%=========================================================
\subsection{Effectiveness of the $\PTM$ Architecture}
%=========================================================
This subsection evaluates the performance of the proposed $\PTM$ architecture. Considering different hyperparameters, such as the number of heads, number of transformer layers, model dimension, and MLP size, we compare different variants of the proposed MTEC architecture in terms of accuracy and the number of parameters (complexity). As it can be seen in Table~\ref{table1}, increasing the number of heads from $8$ to $16$ (Models $1$ and $2$) and model dimension from $32$ to $64$ (Models $1$ and $3$) increase the accuracy of the proposed $\PTM$ framework, while increasing the complexity of the learning model. Similarly, the classification accuracy is improved by increasing the number of transformer layers from $1$ to $2$ (see Models $1$ and $4$) and the MLP size from $256$ to $512$ (Models $3$ and $5$). According to the information provided in Table~\ref{table1}, the best architecture for the proposed $\PTM$ framework is Model $6$. It should be noted that although the number of parameters in Model $6$ is higher than the others, the classification accuracy is higher as well. For that reason, we choose this model to compare it with other state-of-the-art.
%AC: Assuming that model 6 has the higher complexity, a natural question is that why didn't you go further and develop higher complex models, can you also keep Model 4 which has close accuracy with half of the # of parameters.

We also evaluate the effect of window size, which is used for data segmentation during the dataset pre-processing phase. According to Table~\ref{table2},  we consider different window sizes ($9$, $49$, and $99$ days) to build the input samples $\mathcal{D}=\{(\textbf{X}_u, \textbf{y}_u)\}_{u=1}^{M}$. For instance, when the window size is $9$, it means that we use the historical request pattern of content over the past $9$ days to predict the popularity of content in the upcoming time. As shown in Table~\ref{table2}, extending the window length from $9$ to $49$ increases the classification accuracy, while there is no significant improvement considering very large window size, i.e, $99$, due to the degradation in dependency of the number of requests with time.
%AC: can you add some intuition/opinioin on why this happens.

\begin{figure}[t!]
\centering
\includegraphics[scale=0.35]{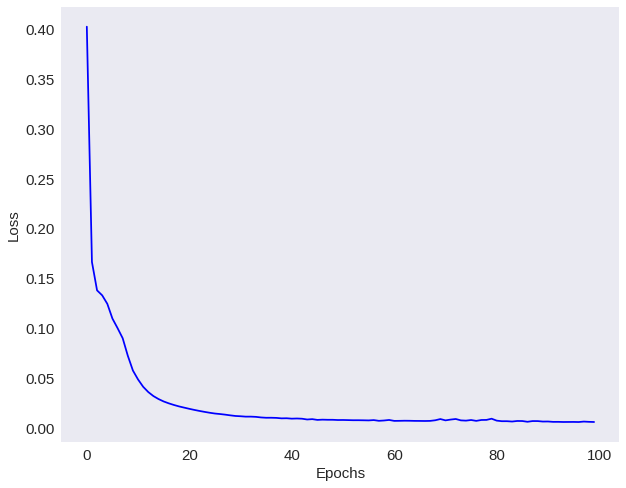}
\caption{The convergence of the proposed $\PTM$ framework.}\label{Loss}
\end{figure}

We evaluate the effect of different loss functions on classification accuracy. According to Eq.~\eqref{loss}, we consider different values for loss weight of each block, denoted by $w_i$, $i=\{1,\ldots,4\}$, to illustrate the effect of loss weight on the overall classification accuracy. Moreover, the convergence of the proposed $\PTM$ framework is illustrated in Fig.~\ref{Loss}. As shown in Fig.~\ref{Loss}, increasing the number of epochs decreases the model loss, which shows that the model is well trained.

%%%%%%%%%%%%%%%%%%%%%%%%%%%%%%%%%%%%%%%%
%%%%%%%%%%%%%%%%%%%%%%%%%%%%%%%%%%%%%%%%

Moreover, to illustrate the superiority of the proposed $\PTM$ architecture in comparison to a single Transformer, we compare it with different Transformer-based networks, such as the single TC model, corresponding to the first path of the $\PTM$ model, the TRC network associated with the second path, the Vision Transformer (ViT) model~\cite{Hajiakhondi2021_ICC}, and the $\PTM$-AT, which is the proposed $\PTM$ architecture with a self-attention layer as the fusion layer instead of the fully connected one.
%AC: Can we use the TC and TRC acronyms to refer to different paths of the model!
According to Table~\ref{table4}, although the complexity of the proposed $\PTM$ network is higher than other architectures, it provides higher accuracy.

%=========================================================
\subsection{Performance Comparisons}
\begin{figure}[t!]
\centering
\includegraphics[scale=0.4]{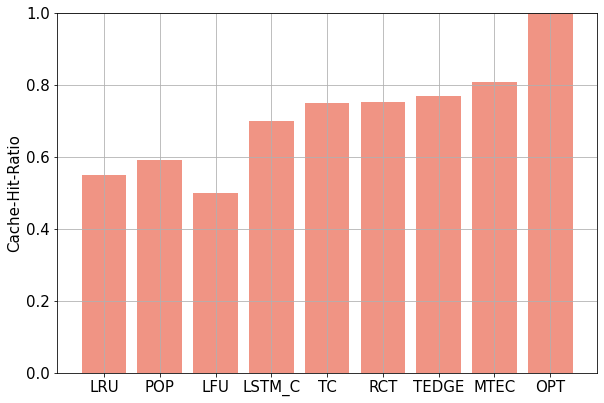}
\caption{A comparison with state-of-the-arts based on the cache-hit ratio. }\label{cachehit}
\end{figure}

%=========================================================
For comparison purposes, we applied seven state-of-the-art caching schemes to the Movielens dataset, including Least Recently Used (LRU)~\cite{Giovanidis2016}, Least Frequently Used (LFU)~\cite{Giovanidis2016}, PopCaching~\cite{Li2016}, LSTM-C~\cite{Zhang2019}, the TC scheme, the RCT scheme, and the TEDGE caching scheme~\cite{Hajiakhondi2021_ICC}, which is based on the ViT architecture.
%AC: Is there another method, not ours, from 2021-2022, that would be good to be included here. If it is time consuming then forget it and we can possibly consider it if it is asked through the revisions.
Fig.~\ref{cachehit} compares the performance of the proposed uncoded $\PTM$ scheme with other baselines listed above in terms of the cache-hit ratio, known as one of the widely used metrics in MEC networks. This metric indicates the number of requests managed by caching nodes versus the total requests made across the network. Note that, the proposed $\PTM$ scheme can be used for both coded/uncoded content placement and the conventional uncoded one. Other baselines, however, are based on uncoded content placement, which is one of the main drawbacks of the existing data-driven caching schemes. In such a case that the multimedia content is partially stored in the storage of caching nodes, the cache-hit ratio would not be an accurate metric for the coded/uncoded content placement framework. For this reason and to be compatible with other state-of-the-arts, we first evaluate the performance of the proposed uncoded $\PTM$ scheme in terms of the cache-hit ratio. Then, we define another metric suitable for the coded/uncoded content placement, known as the transferred byte volume, illustrating the ratio of the data volume, transmitted by caching nodes versus the total volume of the requested contents managed by caching nodes. As shown in Fig.~\ref{cachehit}, the optimal strategy~\cite{Zhang2019} is defined as a caching scheme, where all requests through the network are served by caching nodes, which cannot be obtained in reality. According to the results in Fig.~\ref{cachehit}, the proposed $\PTM$ caching framework achieves the highest cache-hit ratio in comparison to other state-of-the-art counterparts.

\begin{figure}[t!]
\centering
\includegraphics[scale=0.40]{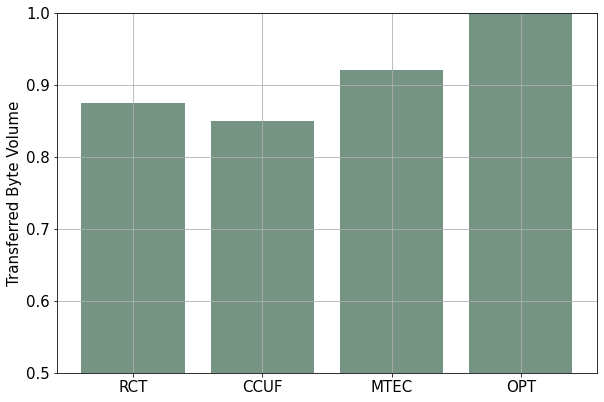}
\caption{A comparison with state-of-the-arts based on the transferred byte volume.}\label{volume}
\end{figure}

In terms of the transferred byte volume, it is assumed that $\alpha = 30$ percent of the storage of FAPs is associated with the popular contents, stored completely, and  $70 \%$ of the storage is assigned to mediocre content, stored partially according to the content placement strategy described in Section~\ref{Sec:2}. As shown in Fig.~\ref{volume}, the byte volume transferred by caching nodes in the proposed $\PTM$ framework is higher than other counterparts. In comparison to the Cluster-centric and Coded UAV-aided Femtocaching (CCUF)~\cite{Hajiakhondi2022}
%AC: Define CCUF
framework, the coded/uncoded content placement in the CCUF is performed based on the historical request probability of content, while the proposed $\PTM$ and RCT frameworks use the predicted one. Moreover, since the classification accuracy of the RCT model is lower than the proposed $\PTM$ architecture, the $\PTM$ framework outperforms in terms of the transferred byte volume.

%%%%%%%%%%%%%%%%%%%%%%%%%%%%%%%%%%%%%%%%
%%%%%%%%%%%%%%%%%%%%%%%%%%%%%%%%%%%%%%%%

%%%%%%%%%%%%%%%%%%%%%%%%%%%%%%%%%%%%%%%%
%%%%%%%%%%%%%%%%%%%%%%%%%%%%%%%%%%%%%%%%

%OOOOOOOOOOOOOOOOOOOOOOOOOOOOOOOOOOOOOOOOOOOOOOOOOOOOOOOO
\section{Conclusion} \label{Sec:5}
%OOOOOOOOOOOOOOOOOOOOOOOOOOOOOOOOOOOOOOOOOOOOOOOOOOOOOOOO
In this paper, we presented an efficient multi-content time-series popularity prediction model referred to as the Multiple-model Transformer-based Edge Caching ($\PTM$) framework with the application to the cluster-centric Mobile Edge Caching (MEC) networks. Due to the lack of predicted request probability, existing data-driven caching strategies were inefficient for coded/uncoded content placement. To tackle this issue, we developed a multiple-model Transformer-based architecture to not only predict the upcoming Top-$K$ popular content but also estimate the request pattern of multiple contents simultaneously, which was used to determine which contents should be stored partially or completely. Simulation results showed that the proposed $\PTM$ caching scheme improves the cache-hit ratio and the transferred byte volume when compared to its state-of-the-art counterparts. A fruitful direction for future research is to implement an integrated Digital Twin (DT)-based popularity prediction model for MEC networks, where mobile users and edge caching nodes cooperatively learn to manage and adapt to changing users' preferences.

%OOOOOOOOOOOOOOOOOOOOOOOOOOOOOOOOOOOOOOOOOOOOOOOOOOOOOOOOO
\bibliography{strings,refs}

\end{document}